\pdfoutput=1
\documentclass[10pt,twocolumn,letterpaper]{article}

\usepackage[pagenumbers]{cvpr} 

\usepackage{graphicx}
\usepackage{amsmath}
\usepackage{amssymb}
\usepackage{booktabs}
\usepackage{multirow}
\usepackage{physics}
\usepackage{diagbox}
\usepackage{enumitem}
\usepackage{xr}

\DeclareMathOperator*{\argmin}{argmin}

\newcommand{\methodName}{GGL}

\usepackage[pagebackref,breaklinks,colorlinks]{hyperref}

\usepackage[capitalize]{cleveref}
\crefname{section}{Sec.}{Secs.}
\Crefname{section}{Section}{Sections}
\Crefname{table}{Table}{Tables}
\crefname{table}{Tab.}{Tabs.}

\def\cvprPaperID{1234} 
\def\confName{CVPR}
\def\confYear{2022}

\begin{document}


\title{Auditing Privacy Defenses in Federated Learning\\ via Generative Gradient Leakage}


\author{Zhuohang Li$^1$ \qquad Jiaxin Zhang$^2$ \qquad Luyang Liu$^3$ \qquad Jian Liu$^1$ \\ $^1$University of Tennessee, Knoxville \qquad $^2$Oak Ridge National Laboratory \qquad $^3$Google Research \\ {\tt\small zli96@vols.utk.edu,} \quad {\tt\small zhangj@ornl.gov,} \quad {\tt\small luyangliu@google.com,} \quad {\tt\small jliu@utk.edu}}


\maketitle

\begin{abstract}

    Federated Learning (FL) framework brings privacy benefits to distributed learning systems by allowing multiple clients to participate in a learning task under the coordination of a central server without exchanging their private data. However, recent studies have revealed that private information can still be leaked through shared gradient information. To further protect user's privacy, several defense mechanisms have been proposed to prevent privacy leakage via gradient information degradation methods, such as using additive noise or gradient compression before sharing it with the server.
    In this work, we validate that the private training data can still be leaked under certain defense settings with a new type of leakage, i.e., Generative Gradient Leakage (\methodName).
    Unlike existing methods that only rely on gradient information to reconstruct data, our method leverages the latent space of generative adversarial networks (GAN) learned from public image datasets as a prior to compensate for the informational loss during gradient degradation.
    To address the nonlinearity caused by the gradient operator and the GAN model, we explore various gradient-free optimization methods (e.g., evolution strategies and Bayesian optimization) and empirically show their superiority in reconstructing high-quality images from gradients compared to gradient-based optimizers.
    We hope the proposed method can serve as a tool for empirically measuring the amount of privacy leakage to facilitate the design of more robust defense mechanisms\footnote{Code is available at: \url{https://github.com/zhuohangli/GGL}}.
\end{abstract}


\section{Introduction}
\label{sec:intro}

Federated Learning (FL)~\cite{mcmahan2017communication,kairouz2019advances,li2020federated} has recently emerged as a new machine learning paradigm that enables multiple clients to collaboratively train a global learning model under the orchestration of a central server. Instead of directly exchanging their private data, each client learns on its local dataset and shares the computed model update or gradient to update the global model. FL places a heavy emphasis on user's data privacy, which has made it particularly suitable for developing machine learning models in privacy-sensitive scenarios such as typing prediction~\cite{hard2018federated}, spoken language understanding~\cite{granqvist2020improving,hard2020training}, medical research~\cite{brisimi2018federated,choudhury2019predicting,sadilek2021privacy}, and financial services~\cite{yang2019ffd,long2020federated}.

\begin{figure}[t]
  \centering
  \includegraphics[width=0.96\linewidth]{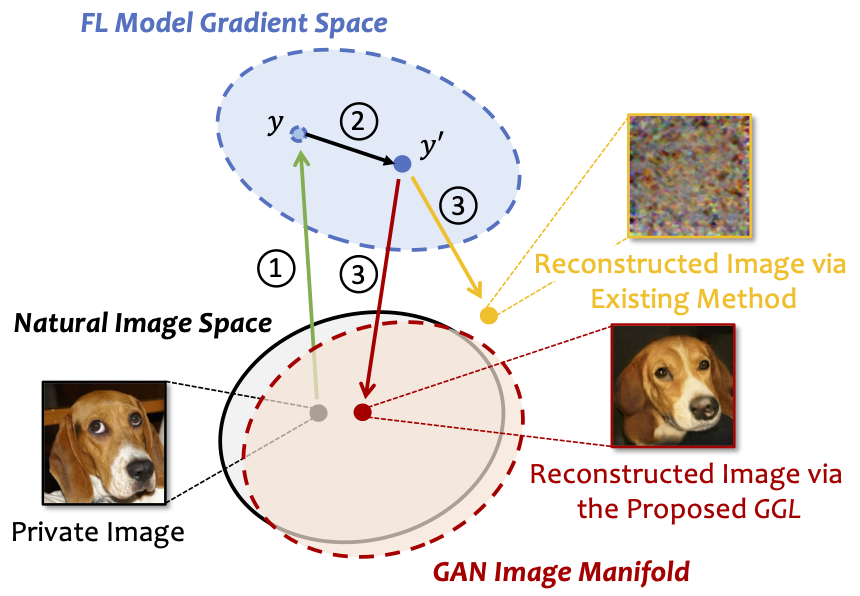}

  \caption{Illustration of data leakage via gradient: \textcircled{1} Client computes gradients on its private data; \textcircled{2} Client applies defense to degrade the computed gradients $y$; \textcircled{3} Adversary attempts to reconstruct the private image from the shared gradients $y'$.}
  \label{fig:overview}

\end{figure}

Although FL is designed to structurally encode data minimization principles to protect privacy,
recent studies have revealed that, in certain cases, sensitive information can still be leaked through the shared gradients~\cite{melis2019exploiting,zhu2020deep,zhao2020idlg,geiping2020inverting,yin2021see}.
To further strengthen FL's privacy properties in these cases,
several defense strategies have been proposed to \textit{degrade} the gradient information before sharing it with the server, such as differential privacy~\cite{geyer2017differentially,wei2021gradient}, gradient compression/sparsification~\cite{zhu2020deep}, and perturbing gradients via data representations~\cite{sun2021soteria}.
These state-of-the-art privacy defenses have been shown to be effective against existing attacks through modifying the gradient information to degrade its fidelity prior to sharing.

A natural question is:
\textit{Can the aforementioned defenses provide sufficient privacy guarantees to prevent the leakage of sensitive information from the client's private data?}
To investigate this, we model the gradient leakage process as an inverse problem, where the goal is to reconstruct the private training data from the client's shared low-fidelity and noisy gradients.
Existing methods seek to solve this inverse problem by iteratively solving for the optimal set of data samples that best match the client's shared gradients via an optimization process (e.g., gradient descent~\cite{geiping2020inverting,yin2021see} or L-BFGS~\cite{zhu2020deep,zhao2020idlg}).
However, such a problem is ill-posed
as there are infinite sets of feasible solutions in the image space and the outcome of the reconstruction may not be a decent natural image. To solve this, existing attacks~\cite{geiping2020inverting,yin2021see} utilize handcrafted image priors such as total variation~\cite{mahendran2015understanding} to regularize the reconstruction process. Although such prior constraint is relatively effective when there is no defense, we find that it is still not sufficiently tight (i.e., many non-image signals can satisfy this constraint) for reconstructing from low-fidelity and noisy gradient observations, causing existing attacks to falsely return unrealistic images when a defense mechanism is applied (e.g., differential privacy), as illustrated in Figure~\ref{fig:overview}.

In this work, we demonstrate on two image datasets that recovering high-fidelity images from shared gradients is still feasible even under certain defense settings by introducing a new type of leakage, namely Generative Gradient Leakage (\methodName).
As shown in Figure~\ref{fig:overview}, our method leverages the manifold of the generative adversarial network (GAN)~\cite{goodfellow2014generative,brock2018large,karras2020analyzing} learned from a large public image dataset as prior information, which provides a good proximation of the natural image space. By minimizing the gradient matching loss in the GAN image manifold, our method can find images that are highly similar to the client's private training data with high quality. However, solving such an optimization problem is not trivial as both the gradient operator and the GAN latent space are highly non-linear and non-convex, and the defense methods applied at the client's side also inject noises into the objective function.
To resolve this, we design an adaptive loss function against common defenses by considering the underlying gradient transformation and resort to gradient-free optimization methods (e.g., evolution strategies~\cite{hansen2016cma} and Bayesian optimization~\cite{eriksson2019scalable}) to search for the global minima within the GAN latent space. We empirically demonstrate that compared with gradient-based optimizers, doing so significantly reduces the chance of converging to a local minimum, leading to a higher quality of reconstructed images as well as improved similarity to the client's private image.
We note that the findings made from the chosen defense settings and datasets may not be general in scope. Nevertheless, we expect the proposed method can serve as a means for privacy auditing in FL by showing how much an adversary can learn under a specific defense setting to assist the future design of privacy mechanisms.

Our main contributions are summarized as follows:
\begin{itemize} [leftmargin=*]

    \vspace{-2mm}
    \item We propose to solve the inverse problem of gradient leakage in FL under noises and defensive transformations by leveraging the prior information learned from deep generative models.

    \vspace{-2mm}
    \item We systematically study $4$ types of gradient-degradation-based defenses, including additive noise, gradient clipping, gradient compression, and representation perturbation, and design adaptive loss functions by accounting for the underlying gradient transformation.
    
    \vspace{-2mm}
    \item To avoid sub-optimal solutions and reveal more private information, we compare different gradient-free optimizers with conventional gradient-based optimizers (e.g., Adam) and experimentally show their superiority for gradient leakage attack in terms of reconstructed image quality and its similarity to the client's private image.

    \vspace{-2mm}
    \item We demonstrate on two image datasets (i.e., CelebA~\cite{liu2015faceattributes} and ImageNet~\cite{deng2009imagenet}) that with the proposed \methodName, high-resolution images can still be recovered from the shared gradients even with the considered defenses, while existing gradient leakage attacks all fail.

\end{itemize}


\section{Related Work}
\label{sec:related}

\subsection{Privacy Leakage via Gradient}

The studies on privacy leakage in FL originate from \textit{membership inference},
where a malicious analyst infers whether a specific data sample has been involved in the training set~\cite{nasr2019comprehensive}.
Moreover, researchers have discovered that the exchanged model updates can be utilized to further infer unintended private information, such as the retrieval of certain \textit{input attributes}~\cite{ganju2018property,melis2019exploiting} (e.g., whether people in the training data wear glasses).
Further studies find it is possible to recover class-level~\cite{hitaj2017deep} or even client-level \textit{data representatives}~\cite{wang2019beyond} (i.e., prototypical samples of the private training set) through generative modeling.

\textbf{Data Reconstruction Attacks.}
Recently, Zhu \textit{et al.}~\cite{zhu2020deep} demonstrate a more severe type of privacy threat where an attacker can fully restore the client's private data samples by solving for the optimal pair of input and label that best matches the exchanged gradients. A follow-up work~\cite{zhao2020idlg} improves on this method by proposing a method for analytically extracting the label information. However, these methods are limited to shallow networks trained with low-resolution images.
A later study by Geiping \textit{et al.}~\cite{geiping2020inverting} extends this attack to more realistic scenarios by successfully restoring ImageNet-level high-resolution data from deeper networks (e.g, ResNet~\cite{he2016deep}) using a magnitude-invariant loss design.
Along this direction, a more recent work by Yin \textit{et al.}~\cite{yin2021see} even achieves image batch reconstruction by utilizing the strong prior encoded in batch normalization statistics.
Despite the improvement, the current research efforts on data reconstruction attacks often assume an ideal setting by targeting a bare-bone FL system without applying any additional privacy-preserving measures or defenses, which contradicts industrial practices.

\subsection{Privacy Preservation in FL}
Existing research efforts for achieving privacy preservation in FL can be generally categorized into \textit{cryptography-based} and \textit{gradient-degradation-based} approaches.

A common type of cryptographic solution is secure multi-party computation (MPC), which aims to have a set of parties to jointly compute the output of a function over their private inputs in a way that only the intended output is revealed to the parties. This can be achieved by designing custom protocols~\cite{mohassel2017secureml,agrawal2019quotient}, or via secure aggregation schemes such as homomorphic encryption~\cite{hardy2017private} and secret sharing~\cite{xu2019verifynet}.
However, merely relying on MPC isn't sufficient to resist inference attacks over
the output~\cite{melis2019exploiting,truex2019hybrid}.

Another line of research seeks to constrain the amount of leaked sensitive information by intentionally sharing degraded gradients. 
Differential privacy (DP) is the standard way to quantify and limit the privacy disclosure about individual users.
DP can be applied at either the server's side (central DP) or the client's side (local DP).
In comparison, local DP provides a better notion of privacy as it does not require the client to trust anyone.
It utilizes a randomized mechanism to distort the gradients before sharing them with the server~\cite{geyer2017differentially,wei2021gradient}.
DP offers a worst-case information theoretic guarantee on how much an adversary can learn from the released data. However, for these worst-case bounds to be most meaningful, they typically involve adding too much noise which often reduces the utility of the trained models.
In addition to DP, it is demonstrated that performing gradient compression/sparsification can also help to prevent information leakage from the gradients~\cite{zhu2020deep}.
A most recent work by Sun \textit{et al.}~\cite{sun2021soteria} identifies the data representation leakage from gradients as the root cause of privacy leakage in FL and proposes a defense named Soteria, which computes the gradients based on perturbed data representations. It is shown that Soteria can achieve a certifiable level of robustness while maintaining good model utility.


\section{Methodology}
\label{sec:method}

\subsection{Threat Model}

\begin{figure}[t]
  \centering
  \begin{subfigure}{0.24\linewidth}
    \centering
    \includegraphics[width=\columnwidth]{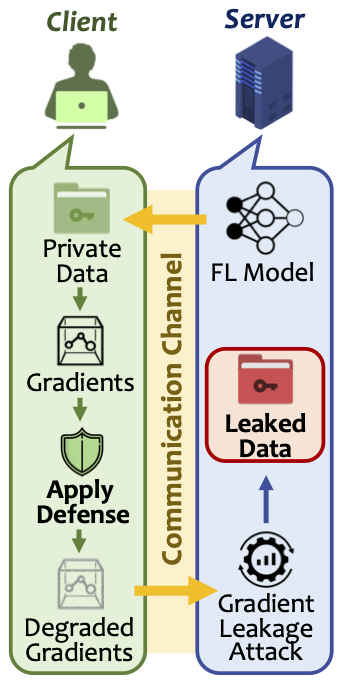}

    \caption{Threat model}
    \label{fig:attack-a}
  \end{subfigure}
  \hfill
  \begin{subfigure}{0.74\linewidth}
    \centering
    \includegraphics[width=\columnwidth]{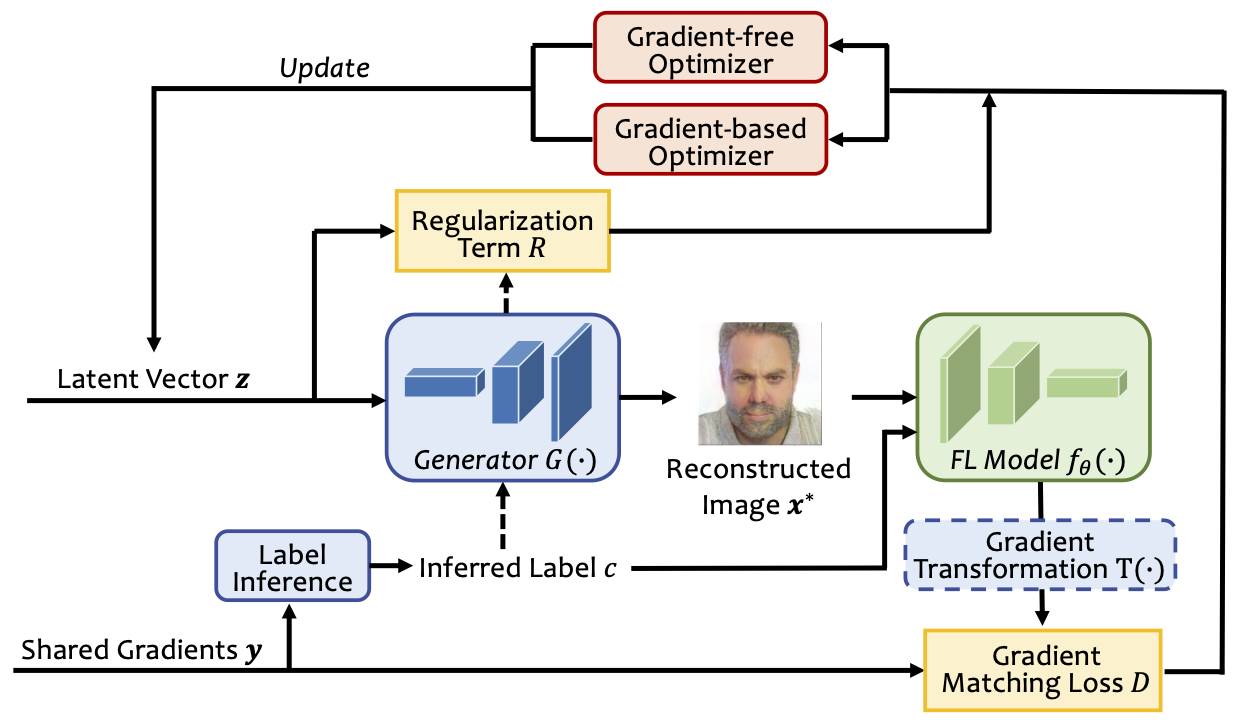}

    \caption{Overview of the proposed method}
    \label{fig:attack-b}
  \end{subfigure}

  \caption{Illustration of the threat model and the proposed method.}

  \label{fig:attack}
\end{figure}

In most existing data leakage attacks~\cite{zhu2020deep,zhao2020idlg,geiping2020inverting,yin2021see}, the adversary is considered to be an honest-but-curious server and has access to the current FL model as well as the shared gradients. As illustrated in Figure~\ref{fig:attack-a}, we further assume that clients apply a privacy defense locally on the gradients computed from their private data, and the adversary can only access the degraded gradients modified by the defense mechanism. The adversary's objective is to reveal as much private information as possible from the degraded gradients. The adversary may or may not know the underlying defense strategy adopted by the client. In either case, the adversary could attempt to launch an adaptive attack by directly using this knowledge or by estimating the defense parameters through the observed gradients.
Additionally, we assume the adversary can utilize the knowledge extracted from publicly available datasets (disjoint from client's private data) to facilitate and improve the attack.

\subsection{Background}
\textbf{Problem Formulation.} 
The task of reconstructing a training image $\vb{x}\in \mathbb{R}^d$ from its gradients $\vb{y}\in\mathbb{R}^m$ can be formulated as a non-linear inverse problem:
\begin{equation}
    \vb{y} = F(\vb{x}), \label{eq:inverse1}
\end{equation}
where $F(\vb{x}) = \nabla_{\boldsymbol\theta} \mathcal{L}(f_{\boldsymbol\theta}(\vb{x}), c)$ is the forward operator that calculates the gradients of the loss $\mathcal{L}$ provided with $\vb{x}$ and its label $c$, along with the FL model $f_{\boldsymbol\theta}$ parameterized by $\boldsymbol\theta$. When defense is applied at the client's side, the reconstructing problem defined in Equation \ref{eq:inverse1} becomes:
\begin{equation}
    \vb{y} = \mathcal{T}(F(\vb{x})) + \boldsymbol\varepsilon,
    \label{eq:problem_defenses}
\end{equation}
where $\mathcal{T}(\cdot)$ is referred to as the lossy transformation (e.g., compression or sparsification) and $\boldsymbol\varepsilon$ means the additive noise (e.g., DP) introduced by the defense algorithm.

\textbf{Current Approach and Its Limitation.}
Existing methods~\cite{zhu2020deep,geiping2020inverting,yin2021see} aim to solve this inverse problem by using image priors in a penalty form:
\begin{equation}
    \vb{x}^* = \argmin_{\vb{x}\in \mathbb{R}^d} \mathcal{D}(\vb{y}, F(\vb{x})) + \lambda\omega(\vb{x}),
\end{equation}
where $\mathcal{D}(\cdot)$ is a distance metric, $\omega(\vb{x}):\mathbb{R}^d \rightarrow \mathbb{R}$ is the standard image prior (e.g., total variance~\cite{beck2009fast} regularization) and $\lambda$ is the weight factor. This form has been demonstrated effective for reconstructing images from the actual gradients. However, when reconstructing from a set of low-fidelity and noisy gradients, such methods would suffer from the limited identification ability of hand-crafted priors, rendering them to return false solutions that are not valid natural images, which is illustrated in Section~\ref{subsec:comparison}.

\subsection{Generative Gradient Leakage}
Motivated by the success of deep generative models for compressed sensing~\cite{bora2017compressed,van2018compressed}, in this work, we aim to leverage a generative model trained on public datasets as the learned natural image prior to ensure the reconstructed image quality.
Moreover, to further account for the privacy defenses that produce degraded gradient information, we propose an adaptive attack by estimating the transformation $\mathcal{T}(\cdot)$ and incorporating it in the optimization process. Specifically, given a well-trained generator $G(\cdot)$, we target to solve the following optimization problem:
\begin{equation}
\label{eq:loss}
    \vb{z}^* = \argmin_{\vb{z}\in \mathbb{R}^k}
    \underbrace{\mathcal{D}\big(\vb{y}, \mathcal{T}\big(F(G(\vb{z}))\big)\big)}_\text{gradient matching loss} + \lambda\underbrace{\mathcal{R}(G;\vb{z})}_\text{regularization},
\end{equation}
where $\vb{z}\in \mathbb{R}^k$ is the latent space of the generative model, $\mathcal{R}(G;\vb{z})$ is a regularization term that penalizes latent vectors which deviate from the prior distribution, and $\lambda$ is the weight factor. Once the optimal solution $\vb{z}^*$ is obtained, the image can be  reconstructed by $G(\vb{z}^*)$. 
An overview of the proposed method is provided in Figure~\ref{fig:attack-b}. We next describe each component in detail.

\textbf{Label Inference.}
Given the shared gradients, the adversary can first adopt an analytical method~\cite{zhao2020idlg} to infer the ground truth label $c$ associated with the client's private image $\vb{x}$.
Specifically, for FL models performing classification task over $n$ classes, the $i^{th}$ entry of the gradients with respect to the weights of the final fully-connected (FC) classification layer (denoted as $\nabla\vb{W}^i_{FC}$) is given by:
\begin{equation}
 \nabla\vb{W}^i_{FC} =\frac{\partial \mathcal{L}(f_{\boldsymbol\theta}(\vb{x}), \vb{c})}{\partial z_i}
\times \frac{\partial z_i}{\partial \vb{W}^i_{FC}},
\end{equation}
where $z_i$ is the $i^{th}$ output of the FC layer.
Note that computing the second term $\frac{\partial z_i}{\partial \vb{W}^i_{FC}}$ results in the post-activation outputs of the previous layer, which will be always non-negative if activation functions like ReLU or sigmoid are applied.
For networks trained with cross-entropy loss on one-hot labels (assuming softmax is applied at the last layer), the first term will be negative if and only if $i=c$.
Thus the ground truth label can be retrieved by identifying the index of the negative entry of $\nabla\vb{W}^i_{FC}$.
The inferred label will be used for evaluating the FL model training loss $\mathcal{L}(f_{\boldsymbol\theta}(\vb{x}), c)$. For conditional GANs~\cite{mirza2014conditional}, the inferred label will also be used as the class condition.

\textbf{Gradient Transformation Estimation.}
The adversary can further attempt to mitigate the impact of the defense by adopting a similar transformation when evaluating the loss of reconstructed images.
Although the transformation process at the client's side isn't directly known to the adversary, the adversary can estimate the parameters of the transformation through the observed gradients.
Specifically, we consider the following defensive transformations (i.e., $\mathcal{T}(\cdot)$):

(1) \textit{Gradient Clipping}: A common technique used in DP studies~\cite{geyer2017differentially,wei2021gradient} to restrict the contribution of each individual client. Given a clipping bound $S$, gradient clipping transforms the gradients as $\mathcal{T}_{cli}(\vb{y}, S)={\vb{y}}/{\text{max}(1, \frac{\norm{\vb{y}}_2}{S})}$. In practice, gradient clipping is often done in a layer-wise manner. The adversary can take the $\ell_2$ norm at each layer of the observed gradients as the estimated clipping bound.

(2) \textit{Gradient Sparsification}: Originally proposed for reducing the communication bandwidth of distributed training~\cite{lin2017deep}, gradient sparsification is also reported to be effective for defending against gradient leakage attacks~\cite{zhu2020deep}. Specifically, given a pruning rate $p\in(0,1)$, the client first computes a threshold $\tau\leftarrow p \text{ of } |\vb{y}|$, which is then used to produce a mask $\mathcal{M} \leftarrow |\vb{y}|>\tau$. Finally, the mask is applied to the gradients during the transformation, i.e., $\mathcal{T}_{spa}(\vb{y}, p)=\vb{y}\odot\mathcal{M}$. This operation is also layer-wise. The adversary can use the percentage of non-zero entries in the observed gradient to estimate its sparsity.

(3) \textit{Representation Perturbation}: The core of the recently proposed Soteria~\cite{sun2021soteria} defense is to prevent data leakage by perturbing the representation learned from a single fully-connected layer $L$ (i.e., the defended layer) to cause maximal reconstruction error. Assume $f_r: \mathbb{R}^d \rightarrow \mathbb{R}^l$ is the feature extractor before the defended layer that maps $\vb{x}\in\mathbb{R}^d$ to a $l$-dimensional data representation $\vb{r}\in\mathbb{R}^l$. Specifically, the client first evaluates the impact of each entry of the representation by computing $\big\{\norm{r_i(\nabla_{\vb{x}}f_r(r_i))^{-1}}_2: i \in\{0, 1, ..., l-1\}\big\}$. Given a pruning rate $p\in(0, 1)$, the client then prunes the $p\times l$ elements in $r$ with the largest $\norm{r_i(\nabla_{\vb{x}}f_r(r_i))^{-1}}_2$ values to get $r'$. Finally, the client computes the gradients on the perturbed representation $r'$. This can be thought as applying a mask only to the gradients of the defended layer: $\mathcal{T}_{rep}(\vb{y}, p)=\vb{y}\odot\mathcal{M}_L$. As this process is deterministic for a given $\vb{x}$ and FL model $f_{\boldsymbol\theta}$, the adversary can reverse-engineer this mask according to the non-zero entries of the gradients from the defended layer.

\textbf{Gradient Matching Loss.}
The first term in the objective function (Equation~\ref{eq:loss}) encourages the solver to find images that are contextually similar to the client's private training images in the generator's latent space  by minimizing the distance between the transformed gradients of the generated images $\vb{\tilde{y}}$ and the observed gradients $\vb{y}$.
We explore the following distance metrics for calculating the gradient matching loss:
(1) \textit{Squared $\ell_2$ norm}~\cite{zhu2020deep,zhao2020idlg,yin2021see}: $\mathcal{D}_1(\vb{y}, \vb{\tilde{y}})=\norm{\vb{y}-\vb{\tilde{y}}}^2_2$; and (2) \textit{Cosine Distance}~\cite{geiping2020inverting}: $\mathcal{D}_2(\vb{y}, \vb{\tilde{y}})=1-\frac{<\vb{y}, \vb{\tilde{y}}>}{\norm{\vb{y}}_2\norm{\vb{\tilde{y}}}_2}$. Cosine distance is magnitude-invariant and is equivalent to optimizing the Euclidean distance of two normalized gradient vectors.

\textbf{Regularization Term.} Optimizing with gradient matching loss alone is likely to produce latent vectors that deviate from the generator's latent distribution, resulting in unrealistic images with significant artifacts. To avoid this issue, we explore the following loss functions to regularize the latent vector during the optimization process: (1) \textit{KL-based regularization}~\cite{kingma2013auto}: $\mathcal{R}_1(G;\vb{z})=-\frac{1}{2}\sum_{i=1}^k \left( 1+\log(\sigma_i^2) - \mu_i^2 - \sigma_i^2 \right)$, where $\mu_i$ and $\sigma_i$ denote the element-wise mean and standard deviation. The KL term aims to reduce the Kullback–Leibler divergence (KLD) between the latent distribution and the standard Gaussian distribution $\mathcal{N}(0, \vb{I})$; and (2) \textit{Norm-based regularization}~\cite{chen2020gan}: $\mathcal{R}_2(G;\vb{z})=(\norm{\vb{z}}^2_2 - k)^2$, which penalizes latent vectors that are far from the prior distribution.

\textbf{Optimization Strategy.}
The target inverse problem described in Equation~\ref{eq:loss} is highly non-linear and non-convex,
and thus choosing the right optimization strategy becomes a critical factor for achieving good image reconstruction.
Existing data reconstruction attacks are all based on gradient-based optimizers such as L-BFGS~\cite{zhu2020deep,zhao2020idlg} and Adam~\cite{geiping2020inverting,yin2021see}. The outcome of such local optimization strategies highly depends on the choice of initialization and often requires multiple trials to find a decent solution. Moreover, we find that for more complex generative models, gradient-based optimizers are likely to converge to local minima, leading to poor reconstruction results.
Inspired by Huh \textit{et al.}~\cite{huh2020transforming},
besides gradient-based optimizers, we further explore two gradient-free optimization strategies to overcome these issues:

(1) \textit{Bayesian Optimization (BO)}~\cite{snoek2012practical}: BO is a global optimization method that can well handle stochastic noise in blackbox functions, which are modeled by a Gaussian process. Vanilla BO scales poorly to high-dimensional problems \cite{snoek2012practical} and thus we adopt a variant of BO, namely, trust region BO (TuRBO)~\cite{eriksson2019scalable}, for performing a global search in the high-dimensional latent space of the GAN model.

(2) \textit{Covariance Matrix Adaptation Evolution Strategy (CMA-ES)}~\cite{hansen2016cma}: CMA-ES leverages a multivariate normal sampling distribution over the search space. At each step, a stochastic search is performed by drawing samples from that distribution to compute the loss. Evolutionary strategies such as recombination and mutation are used to adaptively update its mean and covariance matrix \cite{hansen2006cma}.


\section{Experiments}
\label{sec:exp}

\begin{table}[t]
\centering
\resizebox{\linewidth}{!}{
\begin{tabular}{c@{\extracolsep{6pt}}cccc}
\toprule
\multirow{2}{*}{\diagbox{Reg.}{Grad.}} & \multicolumn{2}{c}{$\mathcal{D}_1$}         & \multicolumn{2}{c}{$\mathcal{D}_2$}          \\ \cline{2-3} \cline{4-5}
                     & MSE-I $\downarrow$           & PSNR $\uparrow$           & MSE-I $\downarrow$            & PSNR $\uparrow$           \\ \toprule
$\mathcal{R}_1$                   & \textbf{0.0320$\pm$0.0173} & \textbf{15.6814$\pm$2.6387} & 0.03671$\pm$0.0227 & 15.3471$\pm$3.1093 \\
$\mathcal{R}_2$                   & 0.0337$\pm$0.0206 & 15.5405$\pm$2.7090  & 0.06290$\pm$0.0815 & 14.3249$\pm$4.1627 \\
\bottomrule
\end{tabular}
}

\caption{Comparison of different loss function configurations.}

\label{tab:loss}
\end{table}

\vspace{-1mm}
\subsection{Experimental Setup}
\textbf{FL Tasks \& Datasets.}
We evaluate our method on two FL tasks: (1) \textit{Gender Classification}: Binary gender classification performed on the CelebFaces attributes dataset (CelebA)~\cite{liu2015faceattributes} with images of size $32\times32$;
and
(2) \textit{Image Classification}: $1000$-class image classification on the ImageNet ILSVRC 2012 dataset~\cite{deng2009imagenet} with images of size $224\times224$.
The FL model for all tasks adopts the ResNet-18~\cite{he2016deep} architecture with randomly initialized weights.
We consider the case where the client performs one local step with batch size $=$$1$ to compute the gradients.

\textbf{Implementation.}
For CelebA dataset, we use the training set containing $162$k images to train a DCGAN~\cite{radford2015unsupervised} on the Wasserstein loss with gradient penalty~\cite{gulrajani2017improved}, while the rest images are reserved for evaluation.
For experiments on ImageNet dataset, we use a pretrained BigGAN~\cite{brock2018large} released by the authors~\cite{biggan_code}.
Note that the FL task is performed on the evaluation set which is disjoint from the GAN training set.
We use the gradients computed from the FL model after applying defenses to conduct reconstruction.

\begin{figure}[t]
    \centering
    \includegraphics[width=\columnwidth]{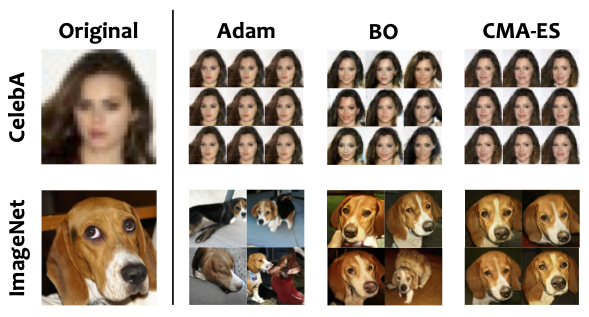}

    \caption{Visual comparison of different optimizers. The images on the right are the reconstruction samples produced by three types of optimizers with different random seeds.}
    \label{fig:optimizer}

\end{figure}

\begin{table}[t]
\centering
\resizebox{\linewidth}{!}{
\begin{tabular}{cc@{\extracolsep{6pt}}cccccc}
\toprule
\multirow{2}{*}{Dataset}  & \multirow{2}{*}{Metric} & \multicolumn{2}{c}{Adam} & \multicolumn{2}{c}{BO} & \multicolumn{2}{c}{CMA-ES} \\ \cline{3-4} \cline{5-6} \cline{7-8}
                          &                         & Mean        & Std.       & Mean       & Std.      & Mean        & Std.      \\ \toprule
\multirow{4}{*}{CelebA}                    & MSE-I $\downarrow$                   & \textbf{0.0427}      & 0.0025     & 0.0813     & 0.0131    & 0.0708      & \textbf{0.0008}    \\
                          & PSNR $\uparrow$                    & \textbf{13.6965}     & 0.2593     & 10.9455    & 0.6816    & 11.4989     & \textbf{0.0533}    \\
                          & LPIPS $\downarrow$                   & \textbf{0.1435}      & \textbf{0.0083}     & 0.2162     & 0.0328    & 0.2136      & 0.0133    \\

                          & MSE-R $\downarrow$                   & \textbf{0.0003}      & \textbf{0.0001}     & 0.0012     & 0.0003    & 0.0015      & 0.0022    \\ \hline
\multirow{4}{*}{ImageNet} & MSE-I $\downarrow$                   & 0.5918      & 0.1955     & \textbf{0.2648}     & 0.0181    & 0.2667      & \textbf{0.0119}    \\
                          & PSNR $\uparrow$                    & 2.4433      & 1.3565     & \textbf{5.7783}     & 0.2992    & 5.7420      & \textbf{0.1988}    \\
                          & LPIPS $\downarrow$                   & 0.7983      & 0.0280     & 0.6166     & 0.0590    & \textbf{0.5736}      & \textbf{0.0209}    \\
                          & MSE-R $\downarrow$                   & 0.1051      & 0.0703     & 0.0035     & 0.0005    & \textbf{0.0018}      & \textbf{0.0002}    \\
                          \bottomrule
\end{tabular}
}

\caption{Quantitative comparison of different optimizers.}

\label{tab:optimizer}
\end{table}

\begin{table*}[t]
\centering
\caption{Quantitative comparison of \methodName~with state-of-the-art methods under various defenses.}

\label{tab:result}
\resizebox{\linewidth}{!}{
\begin{tabular}{cc@{\extracolsep{6pt}}cccc@{\extracolsep{6pt}}cccc@{\extracolsep{6pt}}cccc@{\extracolsep{6pt}}cccc}
\toprule
\multirow{2}{*}{Dataset} & \multirow{2}{*}{Attack} & \multicolumn{4}{c}{Additive Noise~\cite{zhu2020deep,sun2021soteria}}                      & \multicolumn{4}{c}{Gradient Clipping~\cite{geyer2017differentially,wei2021gradient}}                   & \multicolumn{4}{c}{Gradient Sparsification~\cite{zhu2020deep}}                & \multicolumn{4}{c}{Soteria~\cite{sun2021soteria}}        \\ \cline{3-6} \cline{7-10} \cline{11-14} \cline{15-18}
                         &                         & MSE-I $\downarrow$ & PSNR $\uparrow$ & LPIPS $\downarrow$ & MSE-R $\downarrow$ & MSE-I $\downarrow$ & PSNR $\uparrow$ & LPIPS $\downarrow$ & MSE-R $\downarrow$ & MSE-I $\downarrow$ & PSNR $\uparrow$ & LPIPS $\downarrow$ & MSE-R $\downarrow$ & MSE-I $\downarrow$ & PSNR $\uparrow$ & LPIPS $\downarrow$ & MSE-R $\downarrow$ \\
\midrule[\heavyrulewidth]
\multirow{5}{*}{CelebA}  & DLG~\cite{zhu2020deep}                    & 0.6479   & 1.8843  & 0.8197             & 0.0021                            & 0.2097   & 6.7831  & 0.7375             & 0.0326                            & 0.3335   & 4.7679  & 0.7986          & 0.0155                            & 0.3624   & 4.4069  & 0.8007             & 0.0285       \\
                         & iDLG~\cite{zhao2020idlg}                    & 0.6261   & 2.0329  & 0.8209             & 0.0025                            & 0.1960   & 7.0762  & 0.7280             & 0.0326                            & 0.3301   & 4.8124  & 0.8035             & 0.0162                            & 0.3269   & 4.8553  & 0.8036             & 0.0396       \\
                         & IG~\cite{geiping2020inverting}                    & 0.4880   & 3.1151  & 0.8260             & 0.0097                            & \textbf{0.0543}   & \textbf{12.6517}  & 0.2998             & \textbf{0.0003}                            & 0.4103   & 3.8687  & 0.7975             & 0.0113                            & 0.3441   & 4.6326  & 0.8008             & 0.0316       \\
                         & GI~\cite{yin2021see}                   & 0.5738   & 2.4116  & 0.8302             & 0.0023                            & 0.1790   & 7.4701  & 0.7142             & 0.0322                            & 0.2958   & 5.2888  & 0.7775             & 0.0163                            & 0.3179   & 4.9768  & 0.7991             & 0.0409       \\
                         & \textbf{\methodName}                    & \textbf{0.0780}   & \textbf{11.0766}  & \textbf{0.1906}             & \textbf{0.0010}                            & 0.0760   & 11.1902  & \textbf{0.1670}             & 0.0015                            & \textbf{0.0768}   & \textbf{11.1466}  & \textbf{0.1620}             & \textbf{0.0007}                            & \textbf{0.0968}   & \textbf{10.1434}  & \textbf{0.2561}             & \textbf{0.0007}       \\ \hline
\multirow{5}{*}{ImageNet}  & DLG~\cite{zhu2020deep}                     & 0.7438   & 1.2852  & 0.9353             & 0.0049                            & 0.3809   & 4.1912  & 0.9798             & 2.1610                            & 0.4432   & 3.5336  & 0.8907             & 0.0075                            & 0.5990   & 2.2253  & 0.9195             & 0.5415       \\
                         & iDLG~\cite{zhao2020idlg}                    & 0.7352   & 1.3359  & 0.9392             & 0.0041                            & 0.3699   & 4.3190  & 0.9473             & 1.8810                            & 0.4357   & 3.6077  & 0.8935             & 0.0077                            & 0.6089   & 2.1542  & 0.9198             & 0.5425       \\
                         & IG~\cite{geiping2020inverting}                    & 0.3081   & 5.1120  & 0.8677             & 0.4490                            & \textbf{0.1432}   & \textbf{8.4386}  & 0.7476             & 0.0214                            & 0.2993   & 5.2376  & 0.8805             & 0.0501                           & 0.3683   & 4.3373  & 0.8700             & 0.5057       \\
                         & GI~\cite{yin2021see}                   & 0.6593   & 1.8090  & 0.9448             & 0.0031                            & 0.3702   & 4.3154  & 0.9451             & 1.8807                            & 0.4404   & 3.5611  & 0.8889             & 0.0072                            & 0.6235   & 2.0511  & 0.9169             & 0.5792       \\
                         & \textbf{\methodName}                    & \textbf{0.2686}   & \textbf{5.7089}  & \textbf{0.5915}             & \textbf{0.0018}                            & 0.2230   & 6.5163  & \textbf{0.5592}             & \textbf{0.0015}                            & \textbf{0.2141}   & \textbf{6.6920}  & \textbf{0.5170}             & \textbf{0.0017}                            & \textbf{0.2484}   & \textbf{6.0477}  & \textbf{0.5685}             & \textbf{0.0022}       \\
\bottomrule
\end{tabular}
}

\end{table*}

\textbf{Evaluation Metrics.}
Besides qualitative visual comparison, we use the following metrics for quantitative evaluation of the similarity between the target image and the reconstructed image:
(1) \textit{Mean Square Error - Image Space (MSE-I $\downarrow$)}: the pixel-wise MSE between the target image and the reconstructed image;
(2) \textit{Peak Signal-to-Noise Ratio (PSNR $\uparrow$)}: The ratio of the maximum squared pixel fluctuation and the MSE between the target image and the reconstructed image;
(3) \textit{Learned Perceptual Image Patch Similarity (LPIPS $\downarrow$)}~\cite{zhang2018unreasonable}: the perceptual image similarity between the target image and the reconstructed image measured by a VGG network~\cite{simonyan2014very},
and (4) \textit{MSE - Representation Space (MSE-R $\downarrow$)}: the MSE between the target image and the reconstructed image measured in the learned representation space, i.e., the feature vector before the final classification layer~\cite{sun2021soteria}. Note that ``$\downarrow$'' means the lower the metric the higher relative image quality, while ``$\uparrow$'' represents the higher the metric the higher image quality.

\subsection{Choice of Loss Function}
We first evaluate the performance of different loss function configurations.
We randomly select $10$ images from the evaluation set of the CelebA dataset and measure the mean and standard deviation of the MSE-I and PSNR scores between the original images and their reconstructions using Adam optimizer. From results presented in Table~\ref{tab:loss} we observe that using squared $\ell_2$ norm ($\mathcal{D}_1$) for computing the gradient matching loss with KLD as the regularization term ($\mathcal{R}_1$) yields the best reconstructed image quality. Therefore, hereinafter we use this loss configuration for analyzing the impact of different optimizers and defenses.

\subsection{Choice of Optimization Strategy}
We next study the impact of different optimizers on the reconstruction results.
We randomly select images from the CelebA and ImageNet dataset to compute the reconstruction and repeat the experiment by varying its random seed.
The numbers of updates are set to $2500$, $1000$, and $800$ for Adam, BO, and CMA-ES, respectively.
We summarize the results in Table~\ref{tab:optimizer} and provide visualization of the reconstruction samples in Figure~\ref{fig:optimizer}.
We find that the gradient-based and gradient-free optimizers show similar performance on the CelebA dataset, with Adam performing slightly better both visually and statistically. However, on the ImageNet dataset, the gradient-based Adam optimizer fails to recover any useful information from the gradients other than the class label. Moreover, its reconstruction results are highly dependent on the initialization. The gradient-free optimizers (BO and CMA-ES), on the other hand, are still able to find samples that resemble the original private image and are more resilient to different initialization conditions.
The reason causing this performance difference is twofold: (1) the images in the CelebA dataset are well-aligned, while the ImageNet dataset has a more heterogeneous data distribution; and (2) the generator used for generating high-resolution ImageNet data has a deeper and more complex structure, which makes it hard for gradient-based optimizers to find a projection in its latent space.
Based on this observation, we choose to use CMA-ES as the optimizer for conducting experiments under various defense settings.

\begin{figure*}[t]
  \centering
  \begin{subfigure}{0.48\linewidth}
    \centering
    \includegraphics[width=\columnwidth]{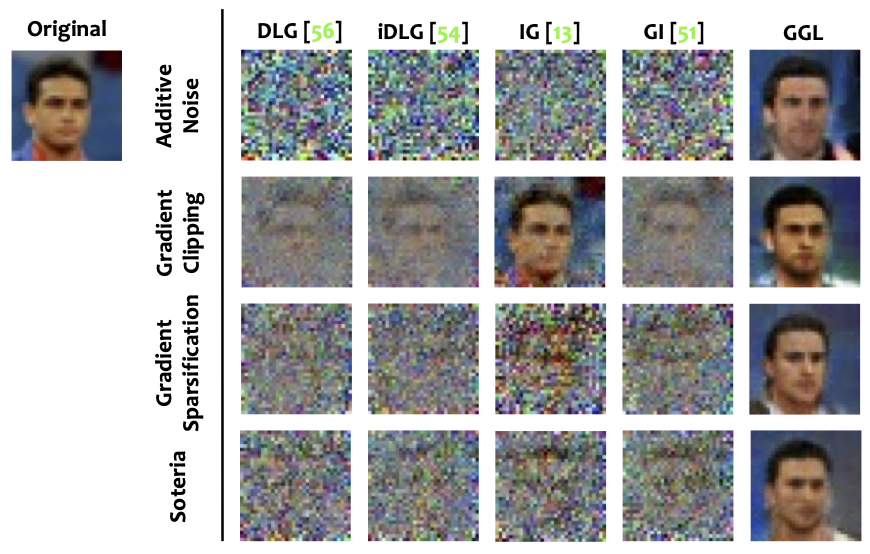}
    \caption{CelebA ($32\times32$ pixels)}
    \label{fig:results-a}
  \end{subfigure}
  \hspace{2mm}
  \begin{subfigure}{0.48\linewidth}
    \centering    \includegraphics[width=\columnwidth]{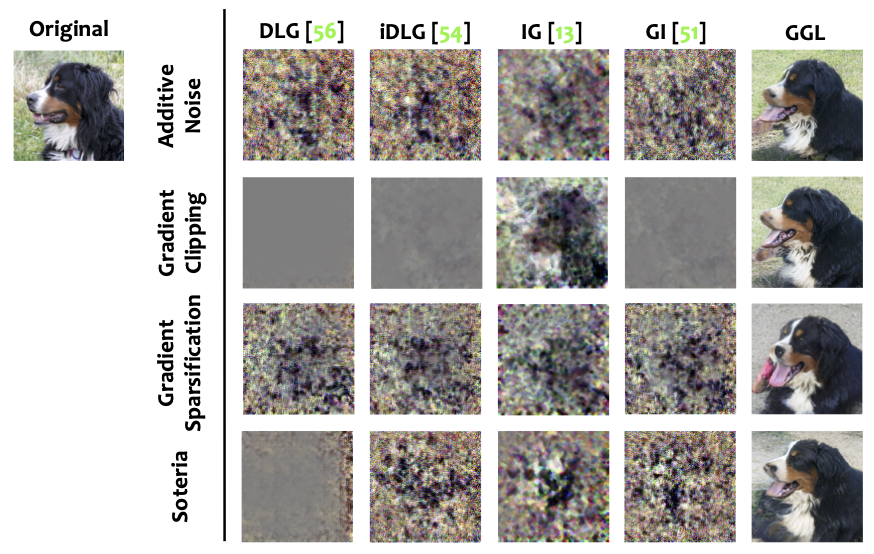}
    \caption{ImageNet ($224\times224$ pixels)}
    \label{fig:results-b}
  \end{subfigure}

  \caption{Comparison of the reconstruction results with attack baselines
  on the CelebA \& ImageNet datasets under various privacy defenses.
  }
  \label{fig:result}

\end{figure*}

\subsection{Comparison with Existing Gradient Leakage Attacks Under Defenses}\label{subsec:comparison}

\textbf{Attack Baselines.}
We compare our method with several state-of-the-art attack methods: 
(1) \textit{Deep Leakage from Gradients (DLG)}~\cite{zhu2020deep}: gradient leakage attack with $\ell_2$ gradient matching loss and L-BFGS optimizer;
(2) \textit{Improved Deep Leakage from Gradients (iDLG)}~\cite{zhao2020idlg}: improved DLG attack with label inference;
(3) \textit{Inverting Gradients (IG)}~\cite{geiping2020inverting}: gradient leakage attack with cosine distance as loss and total variation as prior, optimized using Adam;
and (4) \textit{GradInversion (GI)}~\cite{yin2021see}: gradient leakage attack with $\ell_2$ gradient matching loss and Adam optimizer.

We implemented these attacks following the code repositories released by the authors~\cite{dlg_code,idlg_code,inverting_code}. In our implementation of GI, we consider a stricter scenario where the batch normalization statistics are unknown to the adversary.
For the second-order-based DLG and iDLG attacks, we use the L-BFGS optimizer to conduct $300$ iterations of optimization on the CelebA dataset and $1,200$ iterations on the ImageNet dataset to reconstruct the data.
As for the first-order-based IG and GI attacks, we use the Adam optimizer with an initial learning rate of $0.1$ and conduct $8,000$ iterations of optimization on CelebA and $24,000$ iterations on ImageNet.
The performance of several existing methods is highly varying according to different random seeds.
To mitigate this, each attack is given $4$ trials and the best result with the lowest loss is selected as its final reconstruction.

\textbf{Defense Scheme.}

Following prior studies~\cite{zhu2020deep,sun2021soteria}, we choose a relatively strict defense setting for conducting evaluation:
(1) \textit{Additive Noise}~\cite{zhu2020deep,sun2021soteria}: inject a Gaussian noise $\varepsilon \sim \mathcal{N}(0, \sigma^2\vb{I})$ to the gradients with $\sigma=0.1$;
(2) \textit{Gradient Clipping}~\cite{geyer2017differentially,wei2021gradient}: clip the values of the gradients with a bound of $S=4$;
(3) \textit{Gradient Spasification}~\cite{zhu2020deep}: perform magnitude-based pruning on the gradients to achieve $90\%$ sparsity;
and (4) \textit{Soteria}~\cite{sun2021soteria}: gradients are generated on the perturbed representation with a pruning rate of $80\%$.

\textbf{Results.} Table~\ref{tab:result} compares the performance of the proposed method \methodName~with other gradient leakage attack methods.
Our general observation is that existing attack methods struggle to reconstruct a realistic image with the present of any privacy defense mechanism, while the proposed \methodName~is able to synthesize high quality images that are similar to the original ones, with the measured PSNR $>$$10.1$ on the CelebA dataset and $>$$5.7$ on ImageNet dataset across all scenarios.
One exception is that we find the gradient clipping operation has a very low effect on the IG attack. This is because clipping to $\ell_2$ norm only changes the magnitude of the gradients and does not affect the angular information (i.e., direction). Therefore, though gradient clipping increases the reconstruction error for attacks based on the Euclidean distance between gradients, it will not affect the IG attack which utilizes the magnitude-invariant cosine distance for computing its gradient matching loss. Clipping to $L_\infty$ norm instead would address this issue, however, it is not adopted by existing DP mechanisms as it will result in a poor $\ell_2$ bound.
We also notice that comparing to gradient sparsification, reconstructing from the gradients produced from the perturbed data representation using the Soteria defense would result in higher MSE in both the image space and the representation space, especially on the ImageNet dataset. Despite this, such defense can still be bypassed by our adaptive attack.

\begin{figure}[t]
  \centering
  \includegraphics[width=0.98\linewidth]{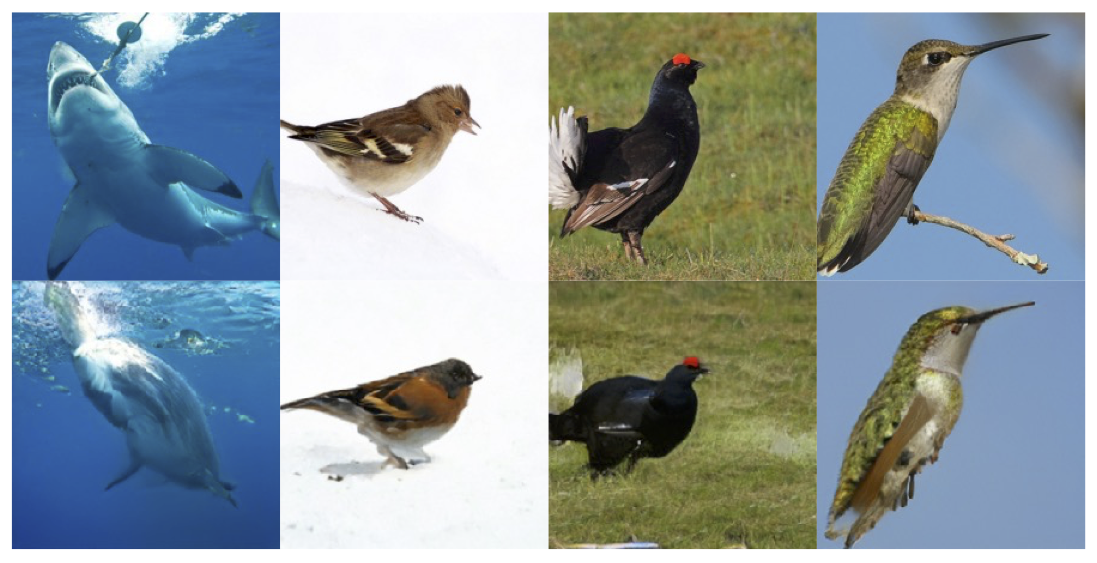}

  \caption{Reconstruction results against the Soteria~\cite{sun2021soteria} defense on the ImageNet dataset: (top) original image and its (bottom) reconstruction by \methodName.}

  \label{fig:samples}
\end{figure}

From the visualization results in Figure~\ref{fig:result}, we can see that except for the IG attack in the case of gradient clipping, the reconstructed image of existing attacks does not reveal much information about the original image.
We also observe that on the CelebA dataset, the proposed method \methodName~isn't able to reconstruct the exact face of the person in the original image when defenses are applied, yet it successfully reveals several key attributes including gender, hair style, hair color, skin color, head posture, and even the background color.
Even on the more challenging ImageNet dataset, our method can still produce a high quality reconstruction that reveals the 
composition of the original image under these defenses.
More samples on the ImageNet dataset against the Soteria defense is presented in Figure~\ref{fig:samples}.

\begin{figure}[t]
  \centering
  \includegraphics[width=0.98\linewidth]{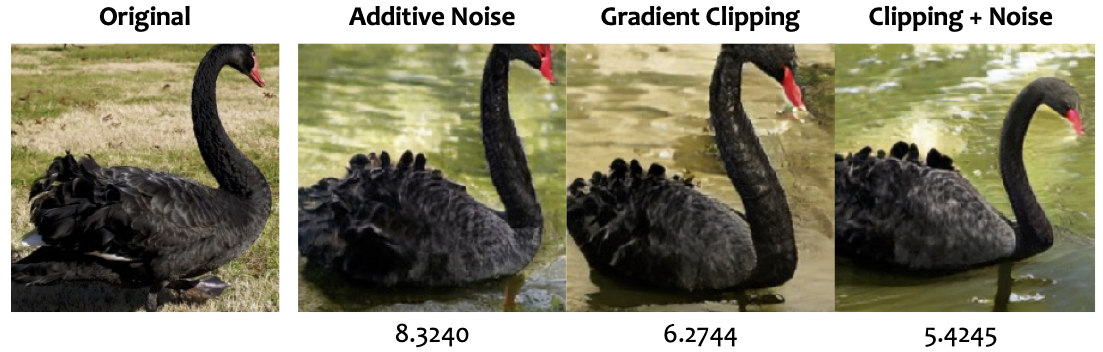}

  \caption{Illustration of combined defense: (left) original image and its (right) reconstruction by \methodName. The PSNR with respect to the original image is shown below each reconstructed image.}

  \label{fig:combined}
\end{figure}

\textbf{Combining Clipping and Noise Addition.} In addition, we also evaluate our attack against the combination of multiple defense mechanisms. Figure~\ref{fig:combined} compares the reconstruction results under $3$ defense settings: additive noise with $\sigma=0.1$, gradient clipping with $S=4$, and simultaneously applying gradient clipping and additive noise (i.e., the privacy defense used in local and distributed DP). We observe that the high-resolution image can still be reconstructed under these defenses, and combining gradient clipping and additive noise would lead to a relatively worse reconstruction with the lowest PSNR. We thus believe this attack can also be used as an auditing measurement for local differential privacy.

\subsection{Impact of Defense Parameter}

\begin{figure}[t]
    \centering
    \includegraphics[width=\columnwidth]{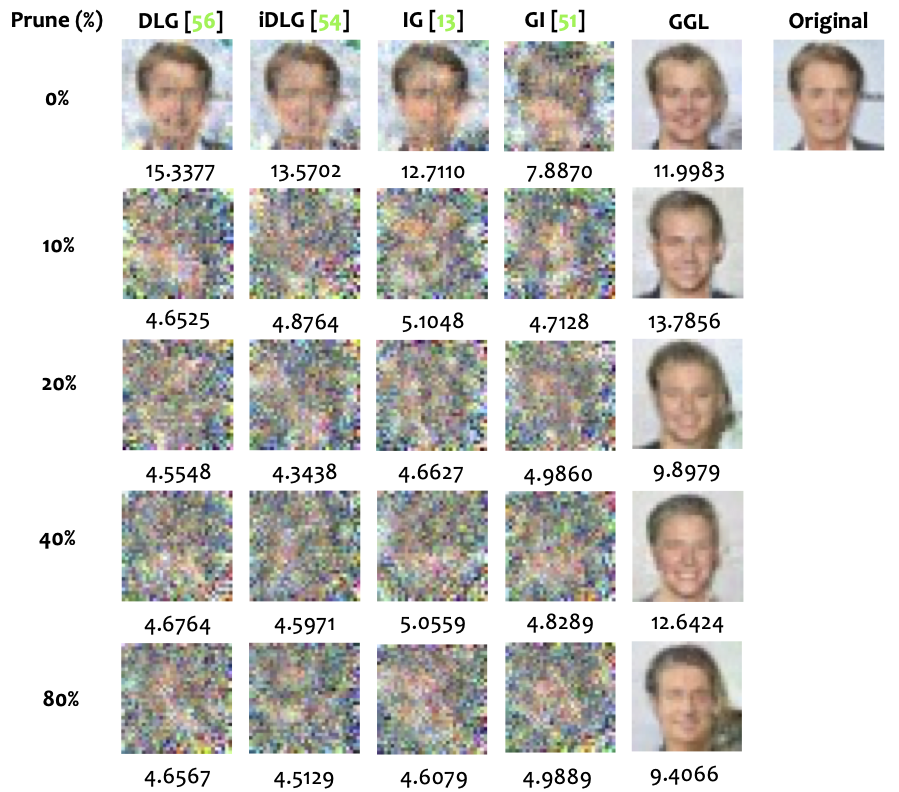}

    \caption{Reconstruction results under the Soteria~\cite{sun2021soteria} defense with varying pruning rates on the CelebA dataset. The PSNR with respect to the original image is shown below each reconstructed image.}

    \label{fig:defense_param}
\end{figure}

We next apply the Soteria~\cite{sun2021soteria} defense on the CelebA dataset as a case study to investigate the impact of different defense parameters.
We use the attack baselines and the proposed \methodName~to generate reconstructions as we vary the pruning rate from $0\%$ to $80\%$, and summarize the results in Figure~\ref{fig:defense_param}.
The authors reported in their original paper~\cite{sun2021soteria} that the DLG~\cite{zhu2020deep} and IG~\cite{geiping2020inverting} attack can tolerate the Soteria defense with a pruning rate up to $40\%$ on the CIFAR10 dataset.
Differently, we observe that on the CelebA dataset, defense with a low pruning rate of $10\%$ would already impose a significant impact on the reconstruction results of these attacks. This is perhaps because the Soteria defense mainly affects the fully-connected layer that produces class-level data representation. Different from CIFAR10, the class-wise label of the CelebA dataset does not directly reveal contextual information about the subject (e.g., the identity of the person). Instead, it only encodes very coarse-grained information (i.e., gender) and thus can be more susceptible to perturbations.
In other words, privacy information that is entangled with the class label is more likely to be leaked through gradients.
Nevertheless, the proposed method can still reliably recover the profile of the person from the remaining gradients regardless of the pruning rate.

\section{Discussion}
\label{sec:discussion}

\begin{figure}[t]
  \centering
  \begin{subfigure}{0.46\linewidth}
    \centering
    \includegraphics[width=\columnwidth]{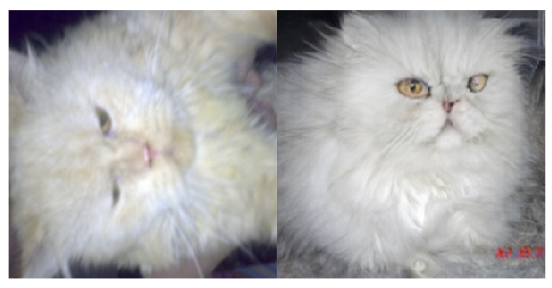}
    \caption{Change in orientation}
    \label{fig:out_of_dist-a}
  \end{subfigure}
  \hspace{2mm}
  \begin{subfigure}{0.46\linewidth}
    \centering
    \includegraphics[width=\columnwidth]{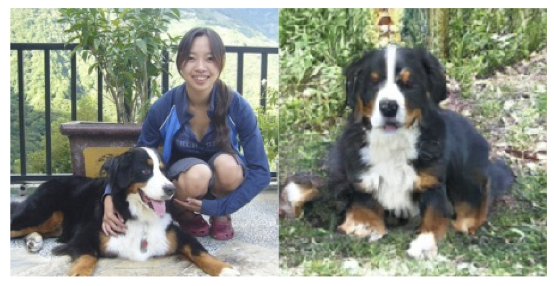}
    \caption{Missing semantics}
    \label{fig:out_of_dist-b}
  \end{subfigure}

  \caption{Reconstruction results of out-of-distribution image samples: (left) original image and its (right) reconstruction by \methodName.}

  \label{fig:out_of_dist}
\end{figure}

\textbf{Limitation.}
Although the image prior captured by the GAN model can help restore the missing information from the degraded gradients for better image reconstruction, at the same time the output image distribution is also constrained by the GAN latent space, rendering it hard to faithfully reconstruct out-of-distribution image samples.
Figure~\ref{fig:out_of_dist} shows two examples of attempting to reconstruct out-of-distribution ImageNet images under the Soteria defense~\cite{sun2021soteria}: in Figure~\ref{fig:out_of_dist-a}, the orientation of the object reconstructed image is changed from the original image; and Figure~\ref{fig:out_of_dist-b}, the reconstruction result is missing important semantics (e.g., the person) that is not well-represented in its class (i.e., Bernese mountain dog).
These phenomena can potentially be improved by jointly optimizing the class condition~\cite{huh2020transforming} or relaxing the generator~\cite{pan2021exploiting}.

\begin{figure}[t]
  \centering
  \begin{subfigure}{0.4\linewidth}
    \centering
    \includegraphics[width=\columnwidth]{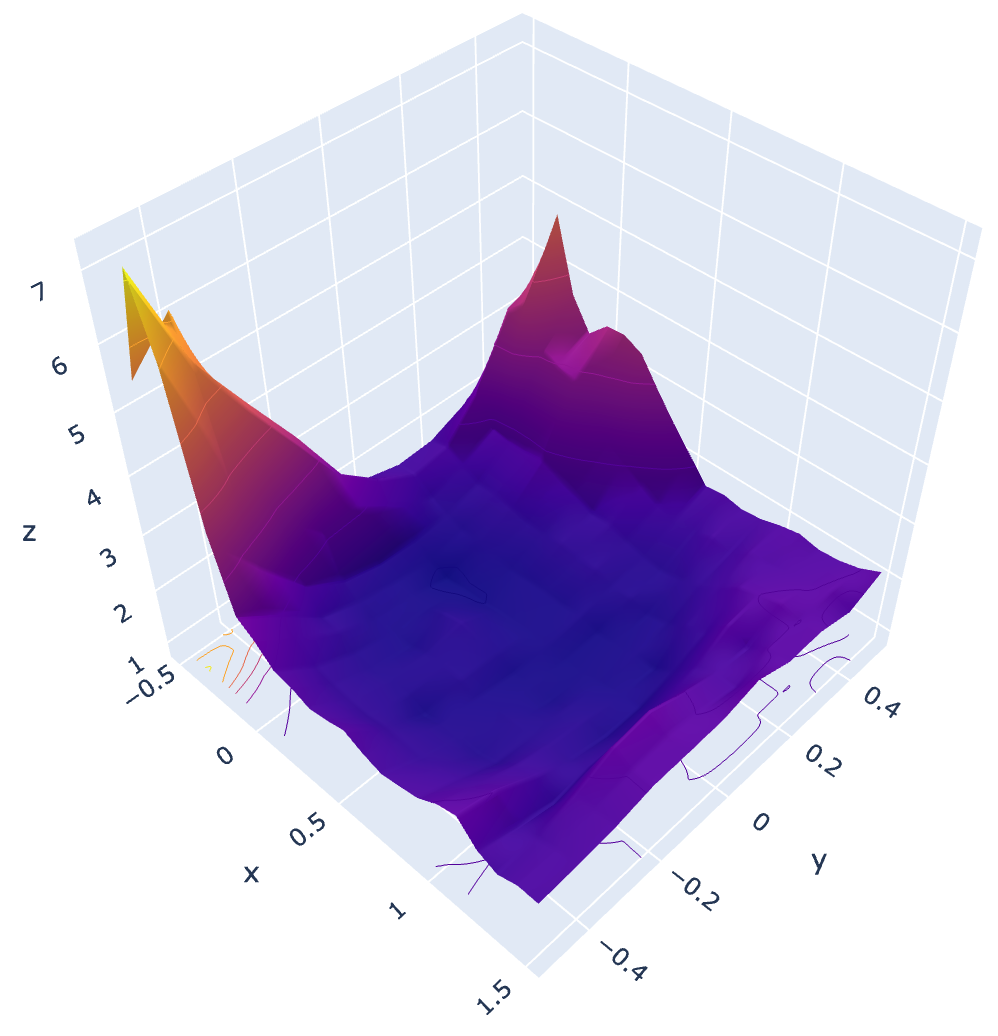}
    \caption{\textbf{Observation}: landscape of the $\ell_2$ gradient matching loss}
    \label{fig:loss-a}
  \end{subfigure}
  \hspace{4mm}
  \begin{subfigure}{0.4\linewidth}
    \centering
    \includegraphics[width=\columnwidth]{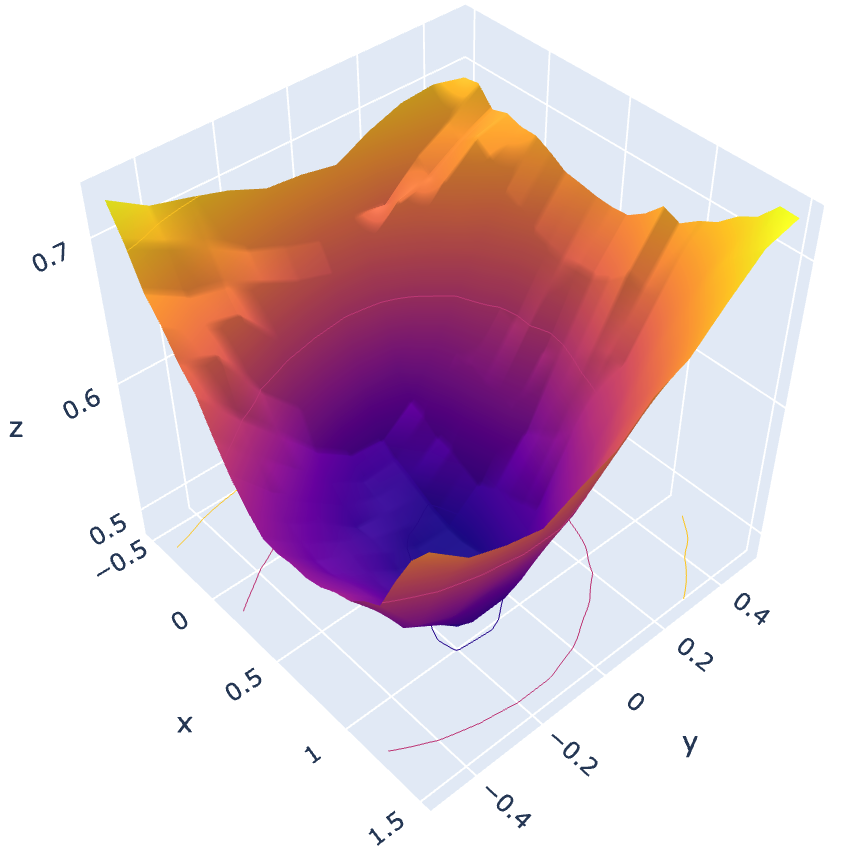}
    \caption{\textbf{Ground truth}: landscape of the LPIPS loss}
    \label{fig:loss-b}
  \end{subfigure}

  \caption{Visualization of the loss landscapes.}

  \label{fig:loss}
\end{figure}

\textbf{Analysis of Loss Landscape and Potential Defense.}
To investigate the reconstruction problem under the constraint of a generative model, we use the latent vector returned by \methodName~as the central point and choose two directions to visualize the loss landscape of the gradient matching loss as well as the LPIPS loss between the original image and the image generated by the BigGAN model by sampling in the latent space. The visualization results are presented in Figure~\ref{fig:loss}, where Figure~\ref{fig:loss-a} shows the loss landscape observed by the adversary if only the gradient information is accessible, and Figure~\ref{fig:loss-b} shows the ground truth loss landscape measured by the LPIPS score assuming the original image is known.
We have the following two observations: (1) the surface of the gradient matching loss is non-convex and contains several local minima; and (2) there exists an inconsistency between the ground truth and the observed loss surface, i.e., the image found by optimizing the gradient matching loss doesn't provide the most similar visual result.
However, as showed in our experiments, such a level of inconsistency isn't sufficient to provide privacy guarantees as the suboptimal result with minimized gradient matching loss still leaks a considerable amount of information about the original image.
This hints us that applying transformations to the gradients to reform the gradient matching loss so that its landscape is no longer in line with the ground truth LPIPS loss can help to effectively achieve privacy preservation against generative gradient leakage attacks.

\section{Conclusion}
\label{sec:conclusion}

This work presents Generative Gradient Leakage (\methodName), an approach that utilizes a generative model to extract prior information from public datasets to improve image reconstruction from degraded gradients produced by privacy defenses.
Our experimental results on two image classification datasets show that with the learned image prior, the proposed method is more resilient to the perturbations and lossy transformations applied to the gradients and is still able to reconstruct high-fidelity images that reveal information about the original images when existing attacks all fail.
We hope the proposed method can serve as an analysis tool for empirical privacy auditing to help facilitate the future design of privacy defenses.

\section*{Acknowledgement} The authors would like to thank Peter Kairouz from Google Research for his valuable feedback on the paper. This work is supported in part
by NSF CNS-2114161, ECCS-2132106, CBET-2130643, the Science Alliance’s StART program, and the GCP credits provided by Google Cloud.
This work is also supported by the U.S. Department of Energy, Office of Science, Office of Advanced Scientific Computing Research, Applied Mathematics program; and by the Artificial Intelligence Initiative at the Oak Ridge National Laboratory (ORNL). ORNL is operated by UT-Battelle, LLC., for the U.S. Department of Energy under Contract DEAC05-00OR22725.




{\small
\bibliographystyle{ieee_fullname}
\bibliography{paper_main}
}

\appendix

\section{Additional Reconstruction Samples}

\begin{figure}[h]
  \centering
  \includegraphics[width=0.98\linewidth]{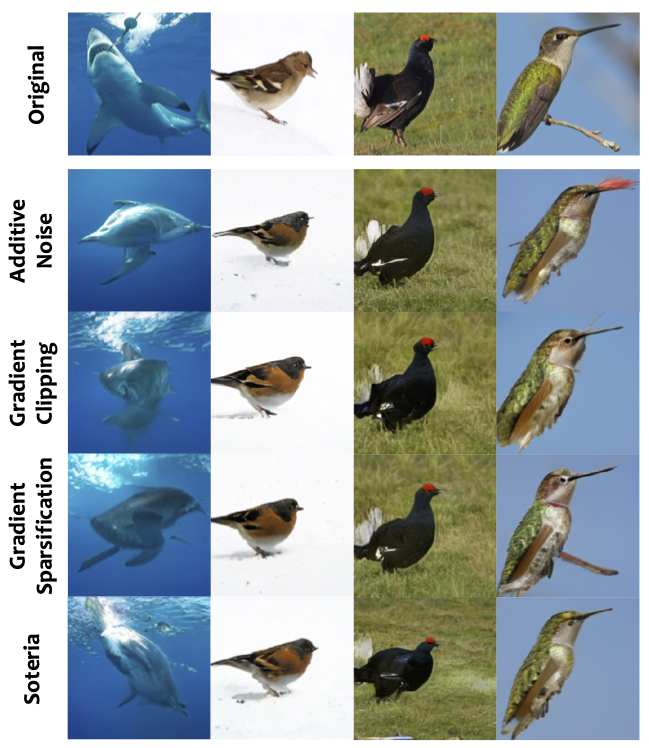}
  \caption{Reconstruction results under various defenses on the ImageNet dataset: (first row) original images and (the rest of rows) their reconstructions by GGL under various defenses.}
  \label{fig:more_sample}

\end{figure}

Due to page limit, we only include the reconstruction results under the Soteria~\cite{sun2021soteria} defense in our main paper (Figure~\ref{fig:samples}) for additional visualization samples on the ImageNet dataset.
Here we present the full results under all $4$ considered defenses (i.e., additive noise~\cite{zhu2020deep,sun2021soteria} with $\sigma=0.1$, gradient clipping~\cite{geyer2017differentially,wei2021gradient} with $S=4$, gradient spasification~\cite{zhu2020deep} with a pruning rate of $90\%$, and Soteria~\cite{sun2021soteria} with a pruning rate of $80\%$) in Figure~\ref{fig:more_sample}.
We observe that our method is able to reconstruct high-quality images from gradients in all these considered cases regardless of the type of defense.

\section{Implementation Details}

\textbf{Optimization Configuration.}
We use the following configuration for the explored optimizers: (1) \textit{Adam}: initial learning rate $lr=0.1$, $\beta_1=0.9$, $\beta_2=0.999$. On the CelebA dataset, we use a step learning rate decay at step $937$, $1562$, and $2189$, by a factor of $\gamma=0.1$. On the ImageNet dataset, the learning rate is linearly warmed-up from $0$ during the first $125$ iterations and gradually reduced to $0$ in the last $625$ iterations using cosine decay;
(2) \textit{BO}: We use the \textit{TurBO-1} algorithm~\cite{eriksson2019scalable} with $256$ initial points, batch size $=10$, lower bound $=-2$, upper bound $=2$, and automatic relevance determination (ARD) kernel for the Gaussian process;
and (3) \textit{CMA-ES}: we use random initialization with batch size $=50$.
We set $\lambda=0.1$ for experiments on the CelebA dataset.
On the ImageNet dataset, for algorithms that do not innately support bound constraints, we apply the \textit{tanh} function to achieve the bound.

\textbf{GAN Configuration.}
For the CelebA dataset, we train a DCGAN~\cite{radford2015unsupervised} with a latent dimension of $128$ with its detailed structure presented in Figure~\ref{fig:GAN_structure}. Specifically, we use the Wasserstein distance with the loss weight set to $10$ for the gradient penalty~\cite{gulrajani2017improved}. The GAN model is trained for $100$ epochs using Adam optimizer with a learning rate of $0.0001$ and a batch size of $64$.
For the ImageNet dataset, we use a pre-trained BigGAN~\cite{brock2018large} with a latent dimension of $128$ and output image size of $256\times256$. The output image is further rescaled to $224\times224$ for computing the FL task.

\begin{figure}[h]
\centering
\begin{subfigure}{0.68\linewidth}
\centering
\resizebox{\linewidth}{!}{
\begin{tabular}{cccc}
    \hline
    Type     & Kernel & Stride & Output \\ \hline
    FC       &        &        & $8192$    \\
    BN1D     &        &        & $8192$    \\
    DeConv2D & $2\times2$    & $2\times2$    & $256$     \\
    BN2D     &        &        & $256$     \\
    DeConv2D & $2\times2$    & $2\times2$    & $128$     \\
    BN2D     &        &        & $128$     \\
    DeConv2D & $2\times2$    & $2\times2$    & $3$       \\ \hline
    \end{tabular}
}
\caption{Generator}
\end{subfigure}
\vspace{2mm}

\begin{subfigure}{0.62\linewidth}
\centering
\resizebox{\linewidth}{!}{
\begin{tabular}{cccc}
\hline
Type   & Kernel & Stride & Output \\ \hline
Conv2D & $3\times3$    & $2\times2$    & $128$     \\
Conv2D & $3\times3$    & $2\times2$    & $256$     \\
Conv2D & $3\times3$    & $2\times2$    & $512$     \\
FC     &        &        & $1$       \\ \hline
\end{tabular}
}
\caption{Discriminator}
\end{subfigure}
\caption{GAN structure for the CelebA dataset.}

\label{fig:GAN_structure}
\end{figure}


\begin{figure}[ht]
  \centering
  \includegraphics[width=0.94\linewidth]{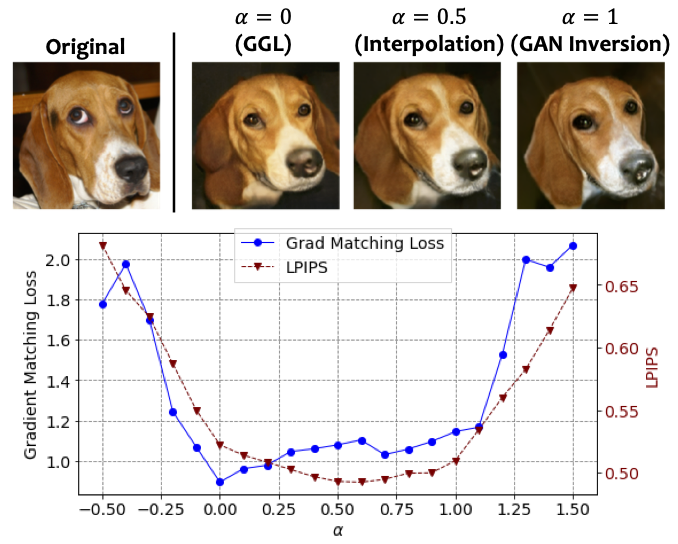}
  \vspace{-2mm}
  \caption{Comparison of image reconstructed by our method and GAN inversion.}
  \label{fig:gan_inversion}
  \vspace{-2mm}
\end{figure}

\section{Loss Landscape Analysis}

\textbf{Comparison with GAN Inversion.}
In our attack, we consider the private image to be unknown and the adversary attempts to reconstruct the image from the shared gradient information using a pre-trained GAN.
However, such reconstruction is constrained by the generator's fitting ability.
GAN inversion technique which inverts a given image to the GAN's latent space can serve as a means for testing the upper bound of the image quality reconstructed from GAN.
To evaluate, we compare the reconstructed image from gradients using our method and the inverted image using GAN inversion technique~\cite{huh2020transforming}.
To compare the information provided by gradient information with the information provided by the original image, we further visualize the gradient matching loss and the LPIPS loss in the GAN latent space. Specifically, we plot the loss functions by interpolating between the latent vectors found by the proposed \methodName~($\vb{z}_1$) and GAN inversion ($\vb{z}_2$): $\vb{z}(\alpha) = (1-\alpha)\vb{z}_1 + \alpha \vb{z}_2$.
From the results presented in Figure~\ref{fig:gan_inversion} we observe that (1) the latent vector found by our method does yield the lowest gradient matching loss on this line; (2) compared to the gradient information, the information provided by the original image can better guide the optimization process in the GAN latent space: the latent vector found by GAN inversion produces a better image quality (lower LPIPS) than the solution found by our method; and (3) the latent vector with the lowest gradient match loss doesn't result in the best image quality/similarity (measured by LPIPS).

\begin{figure}[ht]
  \centering
  \includegraphics[width=0.96\linewidth]{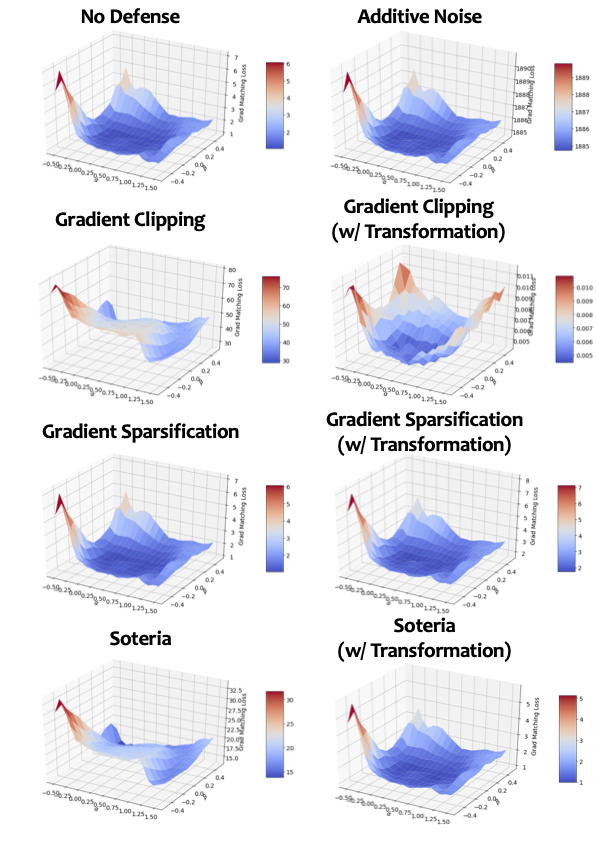}
  \vspace{-5mm}
  \caption{Visualization of observed loss landscapes under various defense settings. The bottom $3$ rows compare the loss surface with (right) and without (left) applying adaptive transformation at the adversary's side.}
  \label{fig:loss_defenses}
  \vspace{-4mm}
\end{figure}

\textbf{Different Defenses.} We next analyze how each defense mechanism affects the loss landscape. We extend the visualization to a 2D surface by adding a second random direction vector $\boldsymbol\eta$ (normalized according to $\vb{z}_2 - \vb{z}_1$): $\vb{z}(\alpha, \beta) = \vb{z}_1 + \alpha(\vb{z}_2 - \vb{z}_1) + \beta \boldsymbol\eta$.
Figure~\ref{fig:loss_defenses} shows the visualized loss surface under different defense settings. We can see that additive noise and gradient sparsification do not have much impact on the geometric landscape of the gradient matching loss, whereas gradient clipping and Soteria~\cite{sun2021soteria} clearly deform the gradient matching loss surface, rendering it hard for the adversary to find a good reconstruction under such defenses. However, by applying the adaptive transformation at the adversary's side, such deformation can be greatly mitigated and thereby enables the adversary to reconstruct high-quality images even with the presence of these defenses.

\section{Larger Batch Sizes or Multiple Local Steps}

Recovering high-resolution batch data with multiple local steps remains a major challenge in this line of research.
Most existing studies~\cite{zhu2020deep,geiping2020inverting} only work on small images (32$\times$32px) for batch size $>$ 1. Currently, the only study that accounts for local steps $>$ 1 is IG~\cite{geiping2020inverting}, but it only works on a single ImageNet image. The only study that can work on batched full-size ImageNet images (224$\times$224px) is GI~\cite{yin2021see}, which supports up to 48 images with local step $=$ 1. However, it can only reveal limited information from partial images of the batch, and it
assumes that the BatchNorm (BN) statistics (mean and std.) of the target batch is jointly provided with the gradients and only works for specially pre-trained large ResNet-50 model (larger model provides more gradient information).

Differently, we seek to investigate the privacy leakage under various defense strategies. We show that even with batch size $=$ 1 and local step size $=$ 1, existing methods still failed to reconstruct the input under defenses, while our method can reveal a good amount of visual information.

\begin{figure}[h]
  \centering

  \includegraphics[width=\linewidth]{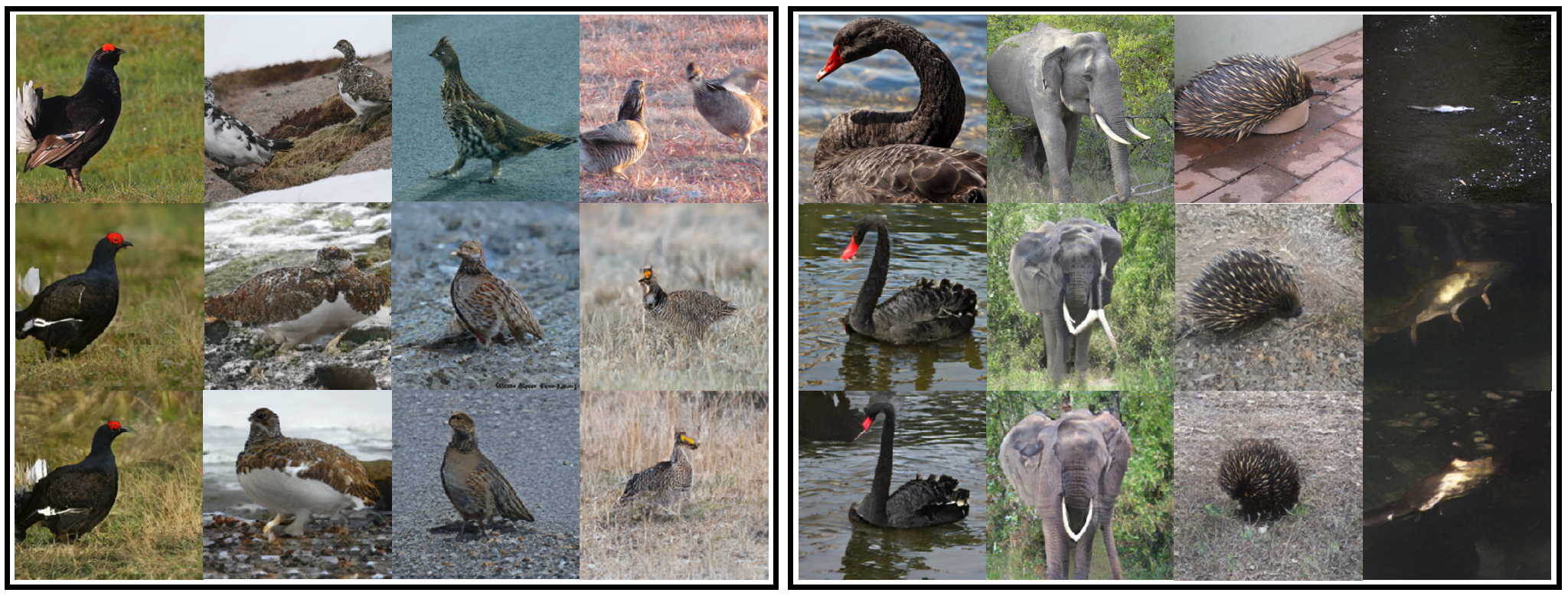}
  \vspace{-6mm}
  \caption{Image reconstruction with batch size $=$ 4: (1st row) original images, (2nd row) reconstructions by GGL w/o defense, and (3rd row) reconstructions by GGL w/ Soteria~\cite{sun2021soteria} defense.}
  \label{fig:batch}
  \vspace{-5mm}
\end{figure}

\begin{figure}[h]
  \centering

  \includegraphics[width=0.99\linewidth]{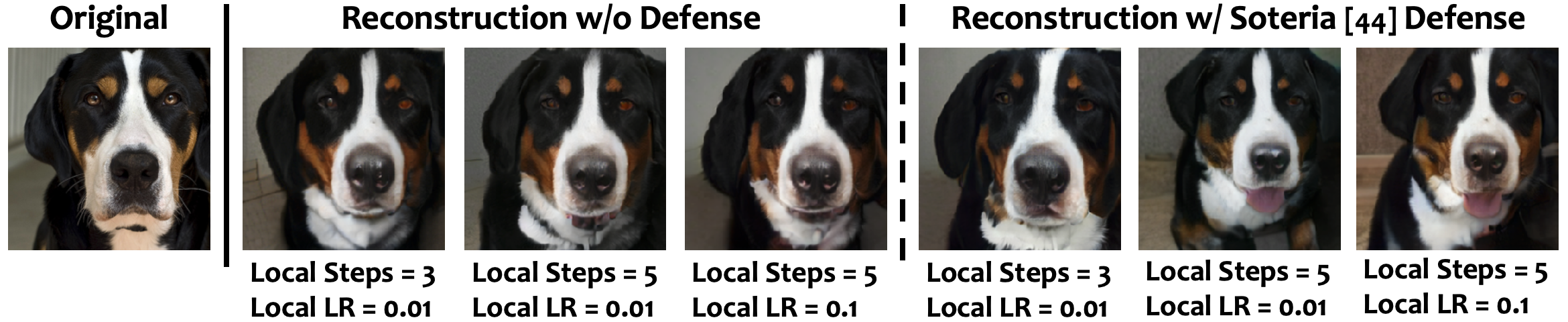}
  \vspace{-2mm}
  \caption{Reconstruction by GGL with multiple local steps.
  }
  \label{fig:step}
  \vspace{-5mm}

\end{figure}

To investigate the generalizability of \methodName, we conducted additional experiments on batched ImageNet images (224$\times$224px) and with multiple local steps,
with the results presented in Figure~\ref{fig:batch} and Figure~\ref{fig:step}, respectively. We can see that \methodName~can still restore a decent amount of visual information under these settings. The proposed \methodName~can be further strengthened with additional prior information (e.g., BN statistics).

\begin{figure}[h]
  \centering

  \includegraphics[width=\linewidth]{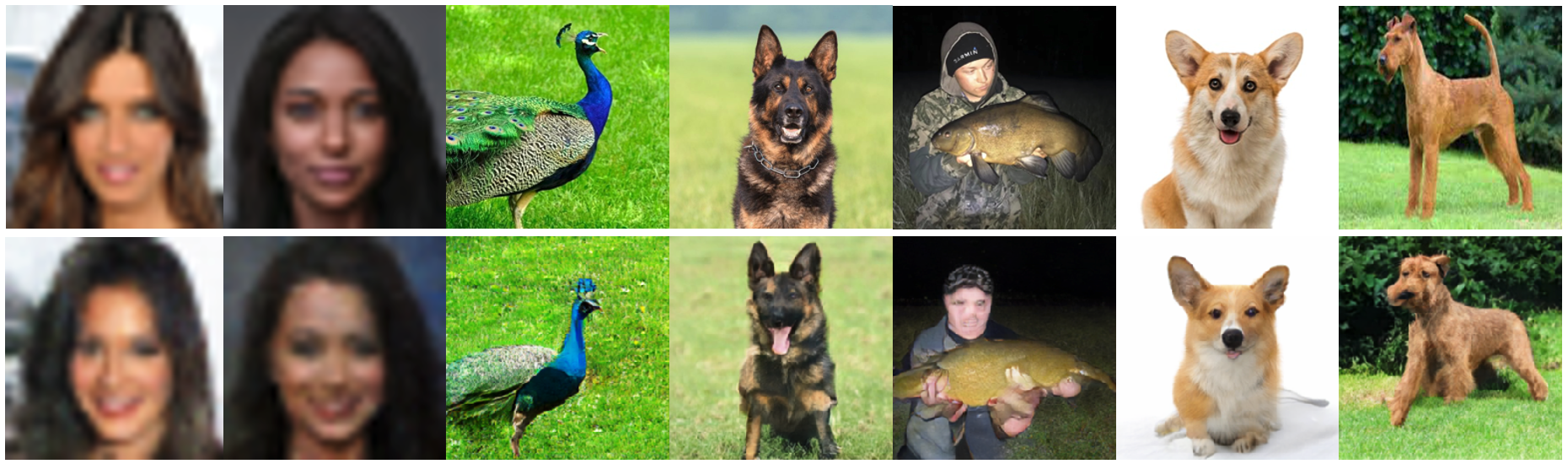}
  \vspace{-5mm}
  \caption{Reconstruction of \textit{in-the-wild} images: (1st row) images from \textit{Google Images} and (2nd row) their reconstructions by GGL.}
  \label{fig:in_the_wild}
  \vspace{-5mm}
\end{figure}

\section{Recovering In-the-wild Data}
We target the practical scenario where the attacker can utilize all public-accessible data as prior information to launch the attack. Thus we chose to use CelebA and ImageNet for evaluation as they are all Internet-based datasets and are easy to access as an attacker. We also used the disjoint dataset so that the images used for testing haven't been used for GAN training.
To investigate the performance of \methodName~under the scenario where the testing image is not from the GAN training distribution,
we conducted additional experiments to recover \textit{in-the-wild} images (i.e., arbitrary images from the search results in Google Images with appropriate cropping/resizing). From the results in Figure~\ref{fig:in_the_wild}, we can see that GGL can still reveal a reasonable amount of visual information even if the testing images are not from the GAN training distribution.

\end{document}